\documentclass[conference]{IEEEtran}
\usepackage{times}

\usepackage[numbers]{natbib}
\usepackage{multicol}
\usepackage[bookmarks=true]{hyperref}
\usepackage{soul}
\usepackage{balance}
\usepackage{graphicx}

\begin{document}

\title{Multimodal Robot Programming by Demonstration:\\ A Preliminary Exploration}

\author{\authorblockN{Gopika Ajaykumar and Chien-Ming Huang}
\authorblockA{
Department of Computer Science\\ 
Johns Hopkins University\\
\texttt{\{gopika,cmhuang\}@cs.jhu.edu}}}

\maketitle


\IEEEpeerreviewmaketitle

\section{Introduction}
Recent years have seen a growth in the number of industrial robots working closely with end-users such as factory workers \cite{international2020world}. This growing use of collaborative robots has been enabled in part due to the availability of end-user robot programming methods that allow users who are not robot programmers to teach robots task actions \cite{ajaykumar2021survey} . \emph{Programming by Demonstration} (PbD) is one such end-user programming method that enables users to bypass the complexities of specifying robot motions using programming languages by instead demonstrating the desired robot behavior \cite{ravichandar2020recent, chernova2014robot}. Demonstrations are often provided by physically guiding the robot through the motions required for a task action in a process known as \emph{kinesthetic teaching}. 

Kinesthetic teaching enables users to directly demonstrate task behaviors in the robot's configuration space, making it a popular end-user robot programming method for collaborative robots known for its low cognitive burden \cite{bambuvssek2019combining, quintero2018robot, ong2020augmented}. However, because kinesthetic teaching  restricts the programmer's teaching to motion demonstrations, it fails to leverage information from other modalities that humans naturally use when providing physical task demonstrations to one other, such as gaze and speech (e.g., \cite{oppenheim2021mental}). Incorporating multimodal information into the traditional kinesthetic programming workflow has the potential to enhance robot learning by highlighting critical aspects of a program \cite{oppenheim2021mental}, reducing ambiguity \cite{saran2020understanding}, and improving situational awareness \cite{penkov2017physical} for the robot learner and can provide insight into the human programmer's intent and difficulties \cite{saran2020understanding}. In this extended abstract, we describe a preliminary study on multimodal kinesthetic demonstrations and future directions for using multimodal demonstrations to enhance robot learning and user programming experiences. 

\begin{figure}[t]
  \includegraphics[width=3.4in]{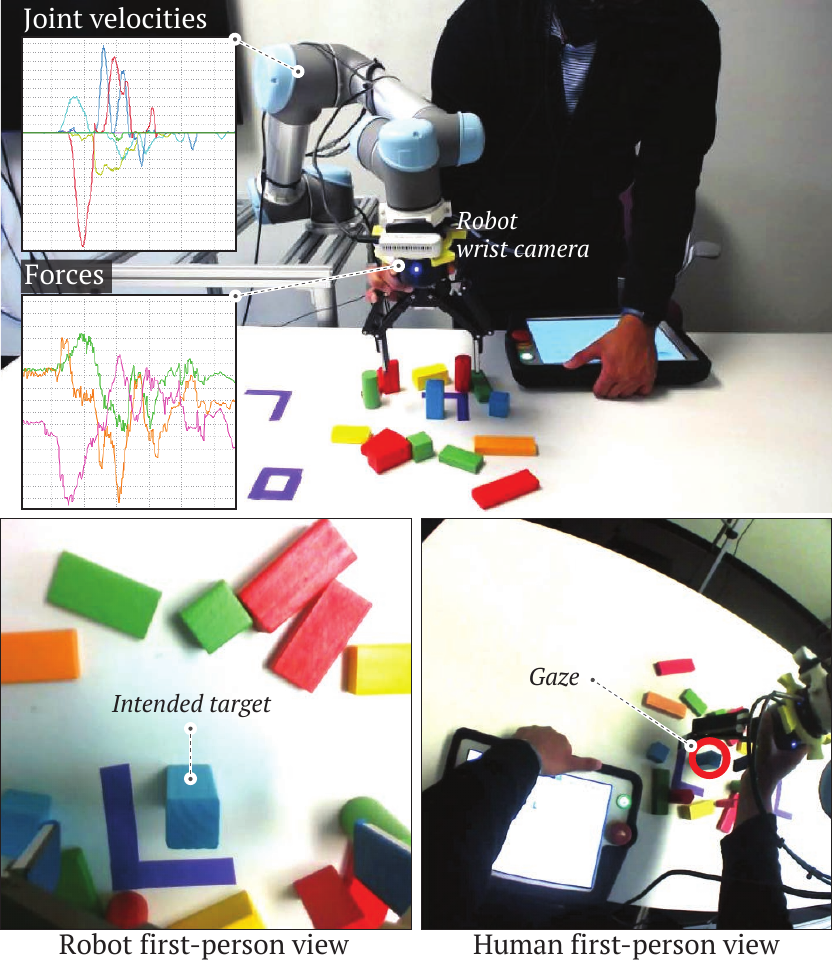}
  \caption{We explore multimodal data from programmers' kinesthetic demonstrations and argue the importance of incorporating natural multimodal cues from human teachers to enable effective robot learning and assisted end-user robot programming.}
  \label{fig:teaser}
\end{figure}

\section{Preliminary Exploration of Multimodal Robot Programming by Demonstration}
\begin{figure*}[t]
  \includegraphics[width=\textwidth]{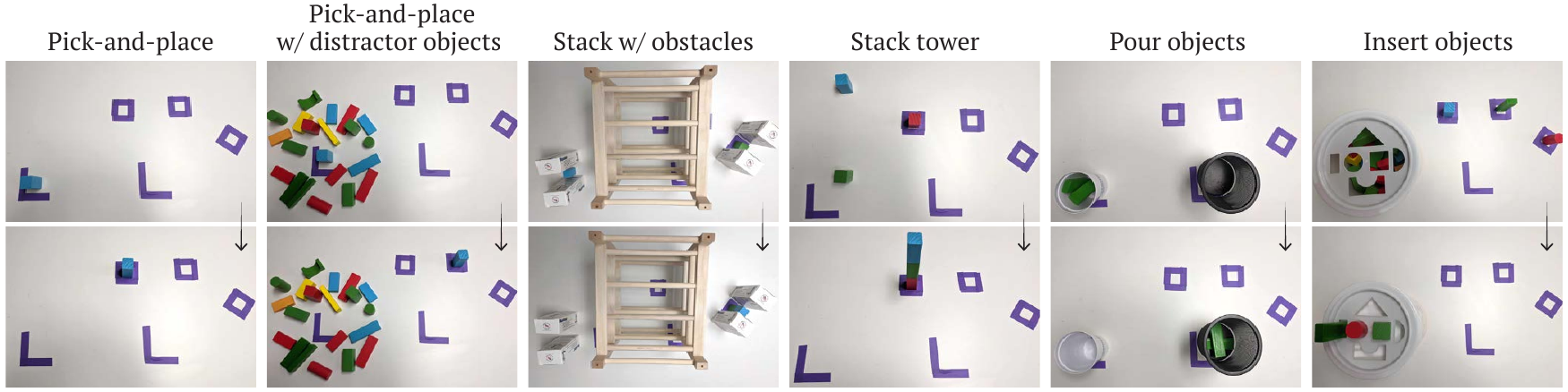}
  \caption{Participants demonstrated six manipulation tasks in our study: pick-and-place, pick-and-place with distractor objects, stack with obstacles, stack tower, pour objects, and insert objects.}
  \label{fig:tasks}
\end{figure*}

We conducted a preliminary exploration of multimodal robot programming by demonstration with 11 users, which included users with and without prior robot programming experience. Participants were instructed to demonstrate six tasks to the robot, which variously involved pick-and-place actions, stacking, pouring, and insertion (Fig. \ref{fig:tasks}), through kinesthetic teaching and by speaking aloud. For each demonstration, we recorded the robot's first-person view collected from its wrist camera and the positions, velocities, and forces of its joints (Fig. \ref{fig:teaser}). In addition, we recorded the user's gaze and first-person view collected using a Pupil Invisible gaze tracker, their speech, and their input forces on the robot's end effector using a force-torque sensor. We also recorded a third-person view of the programming process using a stationary webcam. Below, we describe key observations from our initial exploration of multimodal PbD, with a focus on programmers' narrations during kinesthetic teaching. 



\begin{figure}[b]
\includegraphics[width=3.4in]{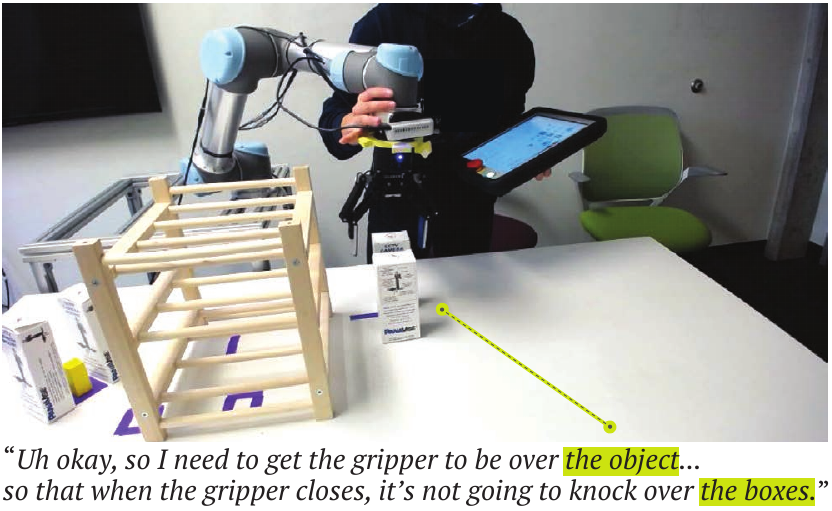}
  \caption{Participants described their actions in terms of objects in the environment when narrating their demonstration, which can enable demonstrated motions to be contextualized within the task environment, including in terms of constraints such as obstacles.}
  \label{fig:verbal-context}
\end{figure}

\subsection{Highlighting Important Aspects for Learning}

Prior work on programming by demonstration has relied on approaches such as clustering multiple task demonstrations (e.g., \cite{akgun2012keyframe}) or segmenting one-shot demonstrations (e.g., \cite{caccavale2019kinesthetic}) to develop skill models that capture the essential robot motions and configurations required for a task. We found that having programmers speak aloud during kinesthetic teaching can help with the process of capturing critical low-level and high-level aspects of a task within a single demonstration. 

\begin{figure}[b]
  \includegraphics[width=3.4in]{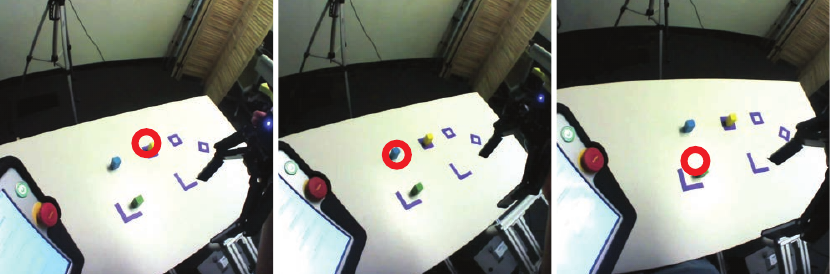}
  \caption{Participants looked at all relevant task objects at the beginning of the demonstration, providing a ``summary'' of the object targets of the demonstrated task. The center of the red circle indicates the participant's gaze.}
  \label{fig:gaze-summary}
\end{figure}

Participants often verbalized how the robot should be positioned with respect to task-relevant objects (e.g., \emph{``I'm going to move the gripper over to the block so the gripper's positioned directly above it''} (P6), \emph{``I'm gonna move the robot over and move the gripper parallel to the table but in the other way''} (P3)), as well as stating the direction (e.g., \emph{``So I'm moving the robot over to the right, um, taking it right above as close as I can''} (P8)) and speed (e.g., \emph{``I'm gonna position the gripper over the block and then slowly close the gripper''} (P6)) of demonstrated motions. In addition, participants occasionally repeated particular phrases in succession to indicate the length of time a motion should continue (e.g., \emph{``close close close close''} (P2) throughout the duration of a grip, \emph{``Keep going. Keep going. Keep going.''} (P5) throughout a continuous trajectory). Participants also indicated key robot configurations within a demonstrated task using phrases such as \emph{``hold here''} (P1) and \emph{``record''} (P11) and described the level of care required for particular motions (e.g., \emph{``Lower the object very carefully''} (P11), \emph{``I need to gently pick it up without hurting the other two paper boxes''} (P10)). Participants' verbal descriptions helped contextualize the motion trajectories and waypoints traditionally used for PbD in terms of the task environment and constraints and may be used to associate demonstration waypoints with reference frames (e.g., \cite{alexandrova2014robot}) (Fig. \ref{fig:verbal-context}). 

In addition to pinpointing aspects of the motion that are important for learning, participants' speech also focused on higher level characteristics of the demonstration related to task order, demonstration goals, and error avoidance. Participants indicated the sequence of actions required for the task (e.g., \emph{``Let it go in a little bit, and then drop it''} (P9)), along with preconditions necessary for any of the task steps (e.g., \emph{``Once it's about in line with the hole I'm going to click open and then try again''} (P8)). Some participants began their demonstration with a summary of the overall task and goal (e.g., \emph{``We're gonna put the figures, the cubes into its corresponding shape, starting with the blue one, and then the yellow, and then the red''} (P7)). Similarly, participants' gaze occasionally ``summarized'' the task in terms of relevant objects at the beginning of the demonstration (Fig. \ref{fig:gaze-summary}). 
Participants also provided warnings about potential errors to avoid throughout their demonstrations, especially related to physical obstacles (e.g., \emph{``Make sure that when the gripper closes it's not gonna hit the table''} (P6), \emph{``Bring the robot back up slowly to avoid colliding with the cardboard boxes''} (P1)). Participants' gaze cues also focused on potential obstacles, which was in line with findings from previous work on using human gaze for assisted teleoperation \cite{aronson2020eye}. Multimodal kinesthetic teaching enables identification of relevant task targets, obstacles, and actions for each portion of a demonstration.

\subsection{Pinpointing Program Suboptimalities}
\begin{figure}[t]
  \includegraphics[width=3.4in]{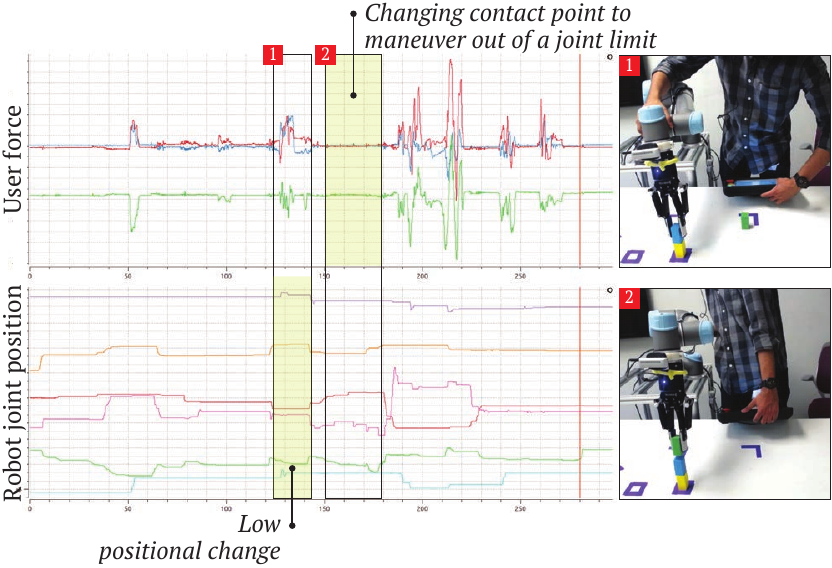}
  \caption{Common motion suboptimalities and difficulties encountered by participants included: (1) encountering a joint limit, which resulted in low positional change in response to user input force and (2) having to change their contact point away from the robot's wrist to maneuver out of a joint limit.}
  \label{fig:motion-difficulties}
\end{figure}

While some demonstration errors correspond with clear signals, such as the robot's emergency stop triggering, others may be more difficult to identify. Multimodal cues can reveal which parts of a programmer's demonstration are suboptimal or erroneous in terms of task efficiency or probability of success. For example, participants often indicated when the robot reached a joint limit during the demonstration (e.g., \emph{``The gripper can't really go down any further''} (P6), \emph{``Moving the elbow is going to be hard. Okay, I think I'm at the max of the joint.''} (P4)), when they can't grip an object from the optimal approach (e.g., \emph{``I'll have to pick it up from, uh, like at an angle''} (P4), \emph{``I will probably just try a bad angle like this''} (P5)), or when they didn't get a good grip on the object with the end effector (e.g., \emph{``Now I'm gonna use the close gripper [command] to grab the block, ah that didn't work so I'm gonna open it.''} (P1), \emph{``Oop, bad grip.''} (P4)). These motion suboptimalities could also be identified by discrepancies between the force input by the user and the resulting positional change of the robot's joints (Fig. \ref{fig:motion-difficulties}). Suboptimal actions in a demonstration often corresponded with participants expressing uncertainty of the actions' success (e.g., \emph{``It's at an angle so I'm not sure if it'll go in.''} (P3), \emph{``If it doesn't stumble, I think it will work now''} (P4)) or laughing. In many cases, participants directly indicated when their demonstration failed or when an error occurred in the course of a demonstration (e.g., \emph{``Oops, I put the arm too far down so I'm gonna move it up a bit''} (P6), \emph{``The angle was a little bit too much so it bounced off position''} (P7)). On the other hand, participants also provided positive feedback when the robot was at a good configuration during the demonstration (e.g., \emph{``That looks like a good position to grab''} (P7)) or when the demonstration was going well by giving quick responses such as ``good'' or ``perfect'' after a task step. Overall, multimodal cues helped distinguish which portions of a demonstration indicated what the robot should do and which did not.

\subsection{Revealing User Challenges and Intent}
\begin{figure*}[t]
  \includegraphics[width=\textwidth]{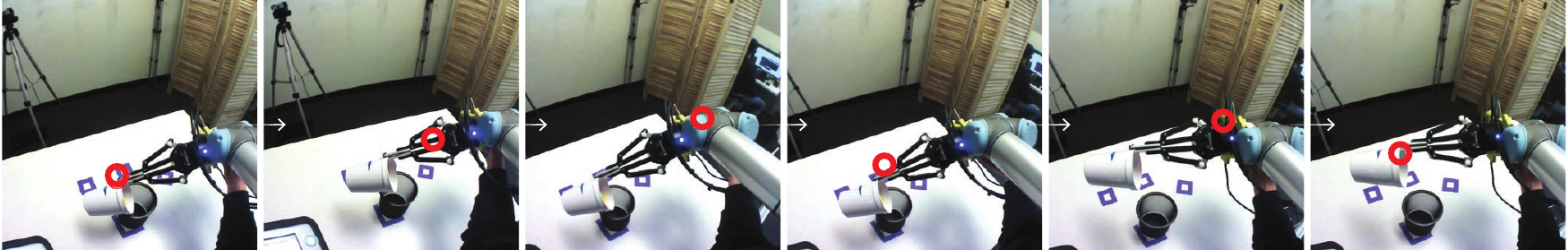}
  \caption{Rapid gaze shifts from robot joints to target objects tended to indicate difficulties with a particular motion such as reaching an optimal end effector alignment.}
  \label{fig:rapid-gaze-shifts}
\end{figure*}

Multimodal cues can indicate challenges users are facing during kinesthetic teaching. Participants frequently indicated when they were having trouble with an aspect of the kinesthetic demonstration (e.g., \emph{``Little hard to find the right angle''} (P6), \emph{``Struggling a little bit with how far the robot can go''} (P4)). Participants also asked questions, directed at themselves, while narrating the demonstration when they encountered a challenge
(e.g., \emph{``How do you even grab this?''} (P9), \emph{``Can I open it? I should open it, right?''} (P4)). When encountering difficulties with a motion, such as reaching the correct alignment or grip, participants' gaze tended to shift rapidly between targets, such as the robot's joint and the task object (Fig. \ref{fig:rapid-gaze-shifts}), which aligns with previous work indicating that users tend to look at problematic joints when experiencing challenges in moving a robot \cite{aronson2020eye}.
Behavioral cues indicating user frustration and fatigue included sighing and frequent shifts of body posture (e.g., bending, changing contact point with robot) (Fig. \ref{fig:motion-difficulties}).

In addition to difficulties, users also described the reasoning behind their actions during kinesthetic teaching (e.g., \emph{``Um, so if I were to do it right now, they would fall off to the side, so I want to move the arm a little bit closer to itself.''} (P7), \emph{``So I'm first moving the robot over and twisting it so that I can close on the edge of the cup''} (P8)). Participants also clarified which portions of a demonstration were allocated towards trial and error or brainstorming purposes (e.g., \emph{``Okay, let's see how we can move these things. Okay, this moves like that. This thing should move this way. Okay, that doesn't tip''} (P9), \emph{``I'm taking a pause to find out the right angle.''} (P4)). Multimodal cues provide insight into programmers' intent and challenges and can signal which parts of a demonstration are intended for human learning (e.g., becoming familiar with the robot's capabilities and constraints) rather than for robot learning (e.g., functional motions). 

\section{Implications of Multimodal Robot Programming by Demonstration}
For robots to fully leverage the rich set of information available in a human teacher's demonstration, they must move beyond solely considering motion aspects of a demonstration. Demonstrations that include the multimodal cues that humans naturally use in teaching, such as gaze and speech, can provide a range of information, from task sequence and goals to programming difficulties, and can pave the way for improved robot learning and easier robot programming.

\subsection{Robot Learning with  Multimodal PbD}
Just as humans make use of multiple modalities such as vision, speech, and sound to obtain a coherent understanding of their environment \cite{ernst2004merging}, robots may obtain a more complete understanding of how a demonstration is situated within a task environment by taking into account multiple modalities in a demonstration, including the human teacher's speech, which can provide semantic information on task objects, and gaze, which can highlight which environmental factors are relevant for the task at hand. Multimodal cues from a human teacher's demonstration can also help robots obtain an understanding of the principal steps involved in a task. In line with prior work indicating humans teach robots using structured processes \cite{ramaraj2021unpacking}, our study participants frequently stated task steps in terms of action preconditions and effects, which can be used by robots for developing task plans automatically based on human demonstrations (e.g., \cite{liang2017framework, pedersen2015gesture}). 

Multimodal PbD can help robots utilize information more effectively within a single demonstration. Multimodal cues can indicate the human demonstrator's focus, which can in turn reveal which aspects of a demonstration are critical to the task being learned (e.g., speed, force, particular configurations, action preconditions, obstacles) and which aspects the robot may have more freedom to stray from (e.g., approach angle for gripping an object). Because multimodal cues can reveal programmer intent and include real-time feedback on a demonstration, they can help robots distinguish between good and failed or suboptimal demonstrations and effectively use a demonstration in its entirety by using portions of a demonstration that correspond to positive teacher feedback to learn what to do and portions of a demonstration that corresponded to negative teacher feedback, program suboptimalities, or programming difficulties to learn what not to do (e.g., \cite{grollman2012robot}). By taking advantage of additional information on how to successfully perform a task and avoid erroneous behaviors from a single demonstration, effective robot learning may occur with a smaller quantity of demonstrations from a human teacher.

\subsection{Assisted Robot Programming with  Multimodal PbD}
Multimodal cues can signal when programmers are encountering difficulties during robot programming. Because interaction modalities such as gaze and speech precede motion during kinesthetic teaching, multimodal behavioral signatures indicating programming difficulties can be identified and difficulties can be prevented early on in a demonstration. Developing programming assistance triggered by multimodal cues, such as autocompletion for pick-and-place tasks or help maneuvering away from a joint limit, may improve the programmer's experience during robot programming and may reduce programmer challenges stemming from the high physical workload involved in kinesthetic teaching (e.g., \cite{quintero2018robot, ajaykumar2020user}). Assistance triggered by the programmer's multimodal data may also help optimize users' demonstrations by reducing the amount of robot motion unrelated to the task at hand within a demonstration (e.g., alignment motions or wrangling the robot into a specific configuration). 

\section{Conclusion and Future Work} 
In this abstract, we presented initial observations into the range of information that narration, gaze, motion, and force data can provide for task learning and assisted robot programming. While prior work has investigated multimodal PbD that takes into account users' motion demonstrations and speech (e.g., \cite{mohseni2019simultaneous}), we believe a multimodal learning approach that takes into account the programmer's gaze and speech and considers motion and force cues indicating user challenges can enable better robot understanding of human demonstrations and more personalized assistance to facilitate easier robot programming by demonstration. Our future work will involve developing computational models that predict when the programmer is experiencing difficulties, such as challenges in maneuvering the gripper to be in line with the target object, based on behavioral signatures such as those observed in our initial study (e.g., Figs. \ref{fig:motion-difficulties} and \ref{fig:rapid-gaze-shifts}). By drawing off of natural multimodal human teaching processes, we aim to minimize the burden of the human teacher in providing optimal demonstrations while expanding the available resources for the robot to understand a teacher's demonstration.


\section*{Ethical Impact Statement}
Modeling users' multimodal cues to improve robot learning and provide online assistance during kinesthetic teaching can benefit users by reducing the user burden in providing large quantities of high-quality demonstrations to the robot learner and in performing complicated maneuvering during kinesthetic teaching. This can help further lower the barriers in robot programming for everyday users of collaborative robots. However, such an approach may also involve risks in the long term. Prior work has suggested that end-users without professional programming experience may be more likely to rely on shortcuts and workarounds during programming \cite{harrison2004editor}. Multimodal data-driven online programming assistance and multimodal robot learning that is robust to program suboptimalities may encourage programmer overreliance on robot learning algorithms and system assistance, possibly encouraging the practice of sloppy programming behaviors from end-users in the long term. Furthermore, online programming assistance based on multimodal cues will involve shifting control over program specification from the human-teacher to the robot learner. While such an assisted programming approach may result in more optimal and robust programs, it could also overly disrupt users' programming workflows or their mental models on how their program works. 

Our goal with this work is to use users' multimodal cues as a means to improve users' experiences with robot programming by minimizing difficulties in kinesthetic teaching. However, we acknowledge that data-driven programming assistance may not always be warranted or desired and could lead to programmer overtrust and automation bias in the long-term. We encourage further work into understanding end-users' perceptions of personalized assistance in real-world scenarios. In particular, future work that builds off of multimodal data-driven robot programming should examine whether users are able to recover from failed online assistance based on misconstrued multimodal cues and whether multimodal programming by demonstration encourages suboptimal teaching from users in the long term.

\section*{Acknowledgments}
This work was supported by the National Science Foundation Graduate Research Fellowship Program under Grant No. DGE-1746891 and the Nursing-Engineering Joint Fellowship from Johns Hopkins University.

\balance
\bibliographystyle{plainnat}
\bibliography{2021-rss-warp-wof-ajaykumar}

\end{document}